\documentclass[sigconf,authorversion,nonacm]{acmart}

\AtBeginDocument{%
  \providecommand\BibTeX{{%
    Bib\TeX}}}

\setcopyright{acmlicensed}
\copyrightyear{2018}
\acmYear{2018}
\acmDOI{XXXXXXX.XXXXXXX}

\acmConference[Conference acronym 'XX]{Make sure to enter the correct
  conference title from your rights confirmation emai}{June 03--05,
  2018}{Woodstock, NY}
\acmISBN{978-1-4503-XXXX-X/18/06}

\begin{document}

\title{DBRouting: Routing End User Queries to Databases for Answerability }

\author{Priyangshu Mandal}
\affiliation{%
  \institution{Indian Institute of Technology Kharagpur}
  \city{Kharagpur}
  \country{India}
}
\email{priyangshu.mandal@kgpian.iitkgp.ac.in}

\author{Manasi Patwardhan}
\affiliation{%
  \institution{TCS Research}
  \city{Pune}
  \country{India}}
\email{manasi.patwardhan@tcs.com}

\author{Mayur Patidar}
\affiliation{%
  \institution{TCS Research}
  \city{Delhi}
  \country{India}
}
\email{patidar.mayur@tcs.com}

\author{Lovekesh Vig}
\affiliation{%
 \institution{TCS Research}
 \city{Delhi}
 \country{India}}
\email{lovekesh.vig@tcs.com}

\renewcommand{\shortauthors}{Mandal et al.}

\begin{abstract}
Enterprise level data is often distributed across multiple sources and identifying the correct set-of data-sources with relevant information for a knowledge request is a fundamental challenge. 
In this work, we define the novel task of routing an end-user query to the appropriate data-source, where the data-sources are databases. We synthesize datasets by extending existing datasets designed for NL-to-SQL semantic parsing. We create baselines on these datasets by using open-source LLMs, using both pre-trained and task specific embeddings fine-tuned using the training data. With these baselines we demonstrate that open-source LLMs perform better than embedding based approach, but suffer from token length limitations. Embedding based approaches benefit from task specific fine-tuning, more so when there is availability of data in terms of database specific questions for training. We further find that the task becomes more difficult (i) with an increase in the number of data-sources, (ii) having data-sources closer in terms of their domains,(iii) having databases without external domain knowledge required to interpret its entities and (iv) with ambiguous and complex queries requiring more fine-grained understanding of the data-sources or logical reasoning for routing to an appropriate source. This calls for the need for developing more sophisticated solutions to better address the task.
\end{abstract}

\begin{CCSXML}
<ccs2012>
<concept>
<concept_id>10002951.10003317.10003338.10003346</concept_id>
<concept_desc>Information systems~Top-k retrieval in databases</concept_desc>
<concept_significance>500</concept_significance>
</concept>
<concept>
<concept_id>10002951.10003317.10003338.10003341</concept_id>
<concept_desc>Information systems~Language models</concept_desc>
<concept_significance>300</concept_significance>
</concept>
</ccs2012>
\end{CCSXML}

\ccsdesc[500]{Information systems~Top-k retrieval in databases}
\ccsdesc[300]{Information systems~Language models}

\keywords{Natural Language Querying on Databases, Query Routing, Enterprise Search}
\begin{teaserfigure}
  \includegraphics[width=\textwidth]{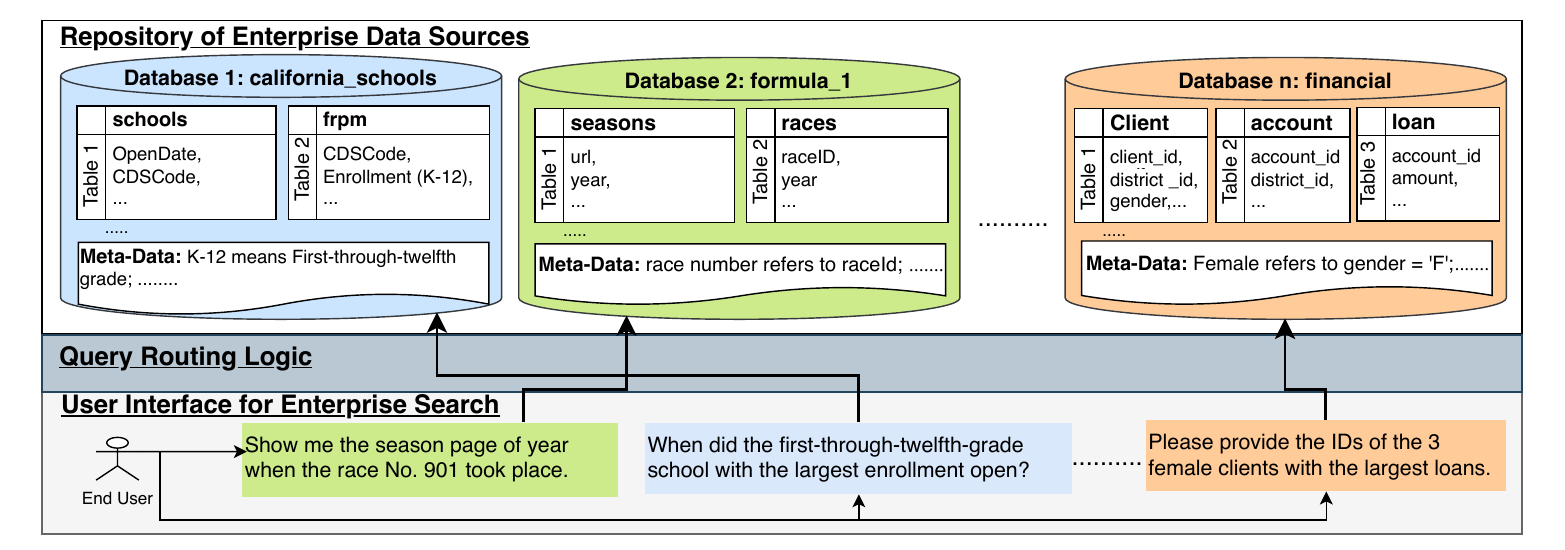}
  \caption{Database (DB) Routing Task}
  \label{fig:dbrouting}
\end{teaserfigure}


\maketitle

\section{Introduction}\label{sec:intro}
Giant enterprises work across multiple verticals (domains) and the relevant data is distributed across multiple data-sources. These sources can be in various forms such as knowledge graphs, databases, document repositories, etc. As a part of an enterprise-level search, when a query in natural language is fired by an end-user, there is a need to route the query to a set of appropriate sources, which carry the data to correctly answer the query, as shown in fig \ref{fig:dbrouting}. Considering this real-life scenario, we define a new task of routing a natural language query to an appropriate data source. We have restricted the data-sources to enterprise databases (DBs) containing information about various domains, which are one of the main types of data-sources in an enterprise setting. 

There is an existing work on routing of natural language (NL) queries to (i) distinct proprietary and open-source Large Language Models (LLMs) \cite{Ong2024RouteLLMLT,Ding2024HybridLC} to strike a balance between performance and cost, (ii) either Retrieval Augmented Generation (RAG) based or Long Context (LC) LLM based approaches \cite{Li2024RetrievalAG}, (iii) distinct Tools or API calls \cite{Qin2023ToolLLMFL,Hao2023ToolkenGPTAF,Patil2023GorillaLL,Schick2023ToolformerLM}. However, to the best of our knowledge, these approaches have not taken into consideration enterprise DBs as data-sources. Also unlike API documentations, which have clear descriptions about API usage, data-sources such as databases may not have high-level descriptions available for their content. Also, for more fine-grained queries such high-level descriptions may not be sufficient. Hence, the approaches applied for Tool/API routing may not be directly applicable to DB routing.

Existing work on question answering over multiple sources mainly take (i) only multiple documents \cite{Schuster2023SEMQASM,Lee2024AmbigDocsRA,Wang2023KnowledgeGP}, (ii) multiple knowledge graphs \cite{Zhang2023TwoIB} or (iii) multiple heterogeneous sources such as knowledge graphs, tables from web and text documents \cite{Christmann2023CompMixAB,Zhang2024SPAGHETTIOQ,Lehmann2024BeyondBA,Zhao2023DIVKNOWQAAT,Christmann2023ExplainableCQ,Sun2018OpenDQ} into consideration.  However, to the best of our knowledge, we are the first ones to define the task of routing queries posed on multiple enterprise DBs. As opposed to other forms of data-sources such as knowledge graphs, text documents, etc, databases have some unique characteristics in that they consist of a combination of structured as well as unstructured data represented in the form of database schema and the associated meta-data or business rules. Moreover, in an enterprise setting multiple databases may contain information from various domains with some domain overlap leading to overlap between DB entities (table names, column names and values), making the routing task even more difficult. In the future, we plan to extend our problem by taking into consideration other data-sources along with DBs for the task of routing and question-answering over heterogeneous data-sources, which would be closer to the realistic scenario of an enterprise search. 

%As opposed to publicly available knowledge sources, such as wikipedia, enterprise databases have unique characteristics in terms of structured format of the knowledge source which is distinct than the text form which is more natural for LLMs to interpret and the need for associated domain knowledge and semantic layer to augment business context to the database entities. 

Due to the unavailability of existing datasets to benchmark for this novel DB routing task, we extend existing datasets designed for NL-to-SQL semantic parsing in the cross-domain setting \cite{yu2018spider,li2024can}. We apply existing embedding and open-source Large Language Model based approaches to benchmark these datasets and also throw some light on the challenges with these approaches. We try to answer the following Research Questions (RQs): 
\begin{enumerate}
    \item Does domain specific (availability of questions from the same database) or cross-domain training assist to improve the performance of the DB-routing task?
    %\item Is the performance of the task when trained with in-domain data (availability of questions from the same database) performance is better than cross-domain?
    \item Does similarity in domains of the data-sources affect the performance of the DB-Routing task?
    \item Does an increase in the number of data-sources affect the performance of the DB Routing Task?
     \item Does domain specific external knowledge facilitate in improving the performance of the DB-Routing task?
\end{enumerate}

The main contributions of this work are as follows:
\begin{itemize}
    \item We are the first ones to define the Database (DB) routing task for routing NL queries to the correct DB.
    \item We extend two existing datasets, viz, Spider \cite{yu2018spider} and BirdSQL \cite{li2024can}, designed for cross-domain semantic parsing task for DB-Routing.
    \item We create benchmarks on these datasets using state-of-the-art embedding, task specific embedding and open-source Large Language Model (LLM) based approaches.
    \item We elaborate on the challenges faced by these approaches for the task, which calls for the need for a more sophisticated solution for the task.
\end{itemize}

\section{Problem Statement}\label{sec:probstmt}
We have an end-user question $q$ and a set of databases $D$ , where each database is indexed by a database id $d_x$ with schema $S_x$ consisting of multiple tables and columns ($T_x$, $C_{xy}$), where $T_x$ are tables that belong to database schema $S_x$ and $C_{xy}$ are the columns of table $T_x$. The task is to rank the databases in $D$ for the question $q$ based on their relevance to the question in terms of answerability (Database can provide correct answer to the question). 

We assume to have training data $Tr = \{Q_{tr},D_{tr}\}$, which provides a mapping for each question $q_{tr} \in Q_{Tr}$ to a database $d_{tr} \in D_{Tr}$. In our current setting, we assume every question has a mapping to atleast one database (all questions are answerable) and is mapped to ONLY ONE database. We assume to have two types of test sets. In-domain test set $Te_{in}$  consists of a set of questions $Q_{in}$ and set of databases $D_{in}$, where $Q_{in} \cap Q_{tr} = \phi$, but $D_{in} = D_{tr}$. This test set is formulated to evaluate the approach with a model trained with the questions posed on seen DBs (hence in-domain).
Cross-domain test set $Te_{out}$ consists of a set of questions $Q_{out}$ and set of databases $D_{out}$, where $Q_{out} \cap Q_{tr} = \phi$ and $D_{in} \cap D_{tr} = \phi$. This test set is formulated to evaluate the approach with a the model trained with the questions of unseen DBs.

We define a scoring function $f(q_i, S_j)$, which takes a question $q_i$ in a test set $Te_{in}$ or $Te_{out}$ and a database schema $S_j$ for a database in a test set $D_{in}$ or $D_{out}$, respectively, as an input and provides a score for that schema for the given question.

% \begin{math}
%     f(q_i, S_j) = P(y_{q,S} = 1| q_i, S_j) \hspace{1cm} \forall S_j of d_j \in  D_{in} or D_{out}
% \end{math}\\

%This score is used to rank the DBs and for selecting the top-k DBs with the highest scores.
The top-k DBs with the highest scores are selected. In some cases, in addition to the schema we take into consideration additional information provided for the DBs, which we call meta-data to compute the score. Meta-data consists of database specific domain knowledge, in terms of column/ column value descriptions, business rules, etc, which may be required to resolve the query. %Meta-data for DB with schema $S_j$ is given as $M_j$. Then the function $f$ becomes:

%\begin{math}
%    f(q_i, S_j, M_j) = P(y_{q,S,M} = 1| q_i, S_j, M_j) \hspace{0.5cm} \forall S_j of d_j \in  D_{in} or D_{out}
%\end{math}\\

\section{Dataset Construction}\label{sec:dataset}
We extend 2 existing datasets constructed for cross-domain NL-to-SQL semantic parsing, viz., Spider \cite{yu2018spider} and BIRD-SQL\cite{li2024can}. A summary of these datasets is provided in Table \ref{tab:dataset_summary}. %In this section we describe these databases and the method we use to construct the datasets for our task.

\subsection{Spider-Route}\label{sec:spider-route}
%The Spider dataset \cite{yu2018spider} consists of train, validation and test sets. The test set is not publicly available. We use the train and validation sets in our experiments. 
Each sample in the Spider \cite{yu2018spider} data set consists of a DB name and a NL question posed on the DB and the corresponding SQL query. For each DB, the information about the table and column names and values is provided. With this information, we define the DB schema using a SQL Data Definition Language (DDL) script (example in Table \ref{tab:schema_format} in Appendix \ref{app:ddl}). 
The train set has 7000 manually annotated samples with questions from 140 databases. The number of questions per database ranges from 7 to 170. The validation set has 1034 questions for 20 databases but the test set is not publicly available. Each DB has 4 to 120 questions and the number of tables per DB is in the range of 2 to 26 and the number of columns per table in the DBs varies from 2 to 48, respectively.

We use this dataset to synthesize datasets for the DB Routing task we name as \emph{Spider-Route}. The questions and databases in the Spider validation set form $Q_{out}$ and $D_{out}$, respectively, to comprise the cross-domain test set $Te_{out}$. The DB on which the question is posed is assumed to be the ground truth DB for that question. We randomly sample $\sim$16\% of questions from each DB in the train set to construct in-domain test set $Te_{in}$, consisting of 140 $D_{in}$ and 1041 $Q_{in}$. We ensure that each of the 140 databases has at least 1 question in the in-domain test set. The original Spider train set after removing the questions belonging to $Q_{in}$ forms our training set $Tr$ consisting of 5959 $Q_{tr}$ and 140 $D_{tr}$. In the original train set, we find some questions are common for multiple databases. For example, the question `Count the number of accounts.' is posed on 3 databases, viz. `cre\_Docs\_and\_Epenses', `small\_bank\_1', `customers\_card\_transactions'. We make sure that such samples stay in the train set so that all questions in the in-domain test set are specific to only one database.

\begin{table}[!t]
  \caption{Summary of Spider-Route and Bird-Route train set}
  \label{tab:dataset_summary}
  \begin{tabular}{l c c}
    \toprule
    Parameter &Spider-Route&Bird-Route\\
    \midrule
    Databases in train set &140&69\\
    Databases in in-domain test set &140&69\\
    Databases in cross-domain test set &20&11\\
    Questions in train set &5959&7948\\
    Questions in in-domain test set  &1041&1480\\
    Questions in cross-domain test set &1034&1533\\
    % Positive samples &5959&15305\\
    % Negative samples &19460&45660\\
    Metadata available &Yes &No\\
    \bottomrule
  \end{tabular}
\end{table}

We manually analyze the databases of each split to form clusters of databases belonging to similar verticals (domains). This helps us to analyze how similarity in domains of the data-sources affects the performance of DB Routing. The manually created clusters of verticals for 20 databases in the \emph{Spider-Route} cross-domain Test set ($Te_{out}$) are illustrated in Table \ref{tab:spider_dev}. 
(clusters of 140 databases of $Te_{in}$ illustrated in Table \ref{tab:spider-in-domain_clusters} in Appendix \ref{sec:verticals}).

To learn task-specific embeddings we train an encoder in a contrastive setting (Detailed in section \ref{sec:meth}) with a train set consisting of positive and negative pairs of a question and a DB schema. We have 5959 positive pairs from $Tr$. To construct the negative pairs, we pair a question with all DBs except the one it originally mapped to, yielding a total of 19460 negative pairs.

\subsection{Bird-Route}\label{sec:bird-route}
We synthesize a dataset for our task we name as \emph{Bird-Route} extending the BirdSQL \cite{li2024can} dataset, which is a collection of DBs from real platforms such as Kaggle, Relation.vit, etc, spanning over diverse professional domains including medical, finance, education, sports, and games. The Bird-SQL dataset contains a train, validation, and a hidden test set. For each question, the dataset provides domain knowledge, termed as 'evidence', required to resolve the query to SQL. Thus, A sample in Bird-SQL consists of a database name, its tables, column names and values, NL question, question-specific evidence, and the corresponding SQL query. Similar to Spider we use this information to define the DB schema. Thus, the BirdSQL consists of a train set with 9428 questions from 69 DBs, whereas the validation set has 1533 questions from 11 DBs. The number of tables and columns per table in the DBs varies between 3 to 13 and 2 to 115, respectively. Thus BirdSQL has large DB schemas as far as token length is considered.

For our task, we assume NO availability of the mapping between the questions and the corresponding evidences. Instead, we assume the availability of DB-level domain knowledge, which we call meta-data, which is a union of all the question-specific evidences (domain statements) for that database. Thus, the number of domain statements per DB varies between 49 to 178. We use the same technique discussed for Spider to create train $Tr$,  $Te_{in}$ and $Te_{out}$, having 7948 questions over 69 DBs, 1480  questions over 69 DBs, and 1533 questions over 11 DBs, respectively. 

As discussed in Section \ref{sec:spider-route}, we tried to create vertical clusters based on the database domains of Bird-Route.  We observe that each of the 11 DBs in \emph{Bird-Route} $Te_{out}$ (DB names provided in Appendix \ref{sec:verticals}) belongs to a distinct vertical, meaning there is no domain overlap between the DBs. The manually formulated vertical clusters of 69 databases of \emph{Bird-Route} $Te_{in}$ are illustrated in Appendix \ref{sec:verticals} in Table \ref{tab:bird-SQL}.

The DB Schema and meta-data do not fit into the context length of the embedding model we use (Detailed in Section \ref{sec:meth}), hence to rank the DBs relevant to a question, we perform the sub-tasks of retrieval of domain statements from the meta-data of each database for a question and retrieval of relevant tables from each database given a question and all relevant domain statements.
To get task-specific embeddings, we train a retriever for the first sub-task (Detailed in Section \ref{sec:fine-tune}), by creating a training set consisting of positive and negative question and domain statement pairs. Positive pairs are constructed using the mapping provided in the original dataset. Hard negative samples are formulated by pairing a question with incorrect domain statements from the ground truth DB (for the question). Soft negative samples are created by pairing a question with domain statements corresponding to a DB distinct from the ground truth DB. In total, we formulate 7398 positive and 19665 negative samples. We also train a retriever for a second sub-task in a contrastive setting. For this, a positive pair is formulated by pairing a question and all question-specific evidences provided in the dataset with a table relevant to the question. We fetch relevant tables for each question using the \textit{From} clause of the ground truth SQL provided in the original dataset. Each relevant table forms a separate positive pair with the question. A negative pair is formed by pairing a question with a string consisting of a table followed by evidence, where the table and/or the evidence belong to the same database as the question but are not relevant to the question. The final dataset for this contrastive training has 15305 positive and 45660 negative samples. %For training a table retriever without the meta-data information (detailed in Section \ref{}), questions are paired with one of its relevant tables without relevant evidences to form positive samples. Questions paired with an irrelevant table of the same database and table of another database form hard and soft negative samples, respectively. This training set has 15305 positive and 27901 negative samples.

%When training with domain knowledgee/evidences we make datasets for 2 models. The first to select the most relevant evidences and the second to find the most relevant tables. The first dataset contains samples of <question, evidence, label>. <question, evidence> pairs in the modified train set form the positive samples. We exclude the samples in the modified train set with no evidence. For negative samples, we fix the number of hard and soft negatives per database. A <question, evidence> pair is hard negative when the evidence belongs to the same database as the question but is irrelevant to the question. It is a soft negative when the evidence is of a different database from the question. In total there are 19665 negative samples and 7398 positive samples. The trained model selects the top-5 evidences in a database for a given question.

\section{Approaches}\label{sec:meth}
We use 3 approaches as the scoring function for the DB ranking problem discussed in Section \ref{sec:probstmt}.
\subsection{Llama3 Language Model (LM)}\label{sec:llm}
We use an instruction-tuned Llama3 70B Language Model with 8K Context Length in a zero-shot setting \cite{Touvron2023LLaMAOA} \footnote{\url{https://huggingface.co/meta-llama/Meta-Llama-3-70B-Instruct}} to rank the databases. We use the prompt illustrated in Table \ref{tab:prompt} in the Appendix \ref{app:prompt}. The prompt contains a list of all DBs with their names, schemas in the Data Definition Language (DDL) Format %, and meta-data associated with those databases, wherever applicable as explained in Section \ref{sec:dataset}
. 
We use the LM to rank the top-K DBs.
For test sets where all the databases do not fit into the token length ($Te_{in}$ of \emph{Spider-Route} and \emph{Bird-Route} and $Te_{out}$ of \emph{Bird-Route}), we tried to use this approach to re-rank the databases for each question in the top-K (10 for \emph{Spider-Route})% and 5 for \emph{Bird-Route}) 
relevant databases with top-K (3 in our case) relevant tables %and top-K (3 for \emph{Bird-Route}) relevant domain statements per database 
retrieved using the cosine similarity of pre-trained sentence-BERT embeddings (Details are provided in Section \ref{sec:pre}). For the \emph{Spider-Route} dataset with DB schemas needing a smaller token length, most of the (10 for $Te_{out}$) databases fit into the allowable 8K token length of  Llama3. However, in the case of \emph{Bird-Route}  very few databases (maximum 5) with very few tables (maximum 3) per database can fit in the context making it infeasible to use Llama3 even for database re-ranking. Hence, we do not use this approach for \emph{Bird-Route}.

\subsection{Pre-trained Embedding Based Similarity}\label{sec:pre}
For each database, we form a textual string consisting of the database name followed by the DDL script of the database schema, consisting of all tables. We find that the token length of this string fits into the allowable token length of the embedding model for the $Te_{out}$ split of \emph{Spider-Route}, whereas for $Te_{in}$ of \emph{Spider-Route} and \emph{Bird-Route} and $Te_{out}$ of \emph{Bird-Route}, it does not.
For $Te_{out}$ of \emph{Spider-Route}, we feed this information to the pre-trained embedding model to get the DB embedding. We use the best performing\footnote{\scriptsize \url{https://www.sbert.net/docs/pretrained_models.html}}  \textit{all-mpnet-base-v2} BERT based model from Sentence Transformers library\footnote{\scriptsize \url{https://huggingface.co/sentence-transformers/all-mpnet-base-v2}} \cite{reimers2019sentence} to compute embeddings with 512 token length. For a given question, we rank the DBs in the repository by using the cosine similarity between their embeddings as a scoring function and choose top-K  (3) DBs.   For other splits, we retrieve the top-K (3) tables relevant to a question by embedding the string formulated by relevant domain statements (in case of \emph{Bird-Route}), followed by the DDL script of each table in the database as text and finding its cosine similarity with the question. For test splits of \emph{Bird-Route},  to retrieve relevant meta-data for a question, we embed each evidence in the meta-data using the pre-trained model and use cosine similarity with the question embedding to retrieve Top-K (3) most similar evidences for that question. To rank the databases we formulate a pooling strategy where we average the cosine similarities of the retrieved top-3 tables with the question to serve as the DB score. The databases are then ranked using this score. 

\subsection{Fine-tuned Task Specific Embeddings}\label{sec:fine-tune}
To learn task-specific embeddings we fine-tune distinct models, using sentence BERT as the base model. For $Te_{out}$ of \emph{Spider-Route} to retrieve DBs relevant to a question, we fine-tune a model to learn embeddings for the questions and the DB schemas. The synthetic data generated for this task using the train split of \emph{spider-route}, is explained in Section \ref{sec:spider-route}.  For $Te_{in}$ of \emph{Spider-Route} and \emph{Bird-Route} and $te_{out}$ of \emph{Bird-Route}, to facilitate the retrieval of appropriate tables in a DB for a given question (in context of relevant metadata in case of \emph{Bird-Route})
we fine-tune a model (one for each dataset) to learn embeddings for question and database tables.  For \emph{Bird-Route} dataset, to facilitate the retrieval of appropriate domain statements from DB meta-data, we fine-tune a model to learn embeddings for both the question and domain statements.
The synthetic data generated for these tasks, using the train split of \emph{Bird-Route},  is explained in Section \ref{sec:bird-route}.
 
We fine-tune the sentence BERT model with a constrastive loss to create task-specific embeddings for the above tasks. The training samples needed for this loss are in the form <sentence 1 $s_i$, sentence 2 $s_j$, label $l$>, where the `label' decides if the sentence embeddings are to be brought closer in the embedding space (when the label is 1) or pushed apart (when the label is 0). The loss function is given by:
\begin{math}
    Loss(z_i,z_j) = 0.5* (l* cos\_sim(z_i,z_j)^2 + (1-l)* ReLU( m - cos\_sim(z_i, z_j)^2))
\end{math}
Where, $z_i$ and $z_j$ are the embeddings of the $s_i$ and $s_j$, i.e. $z_i$ = $f(s_i)$ and $z_j$ = $f(s_j)$. $f(.)$ is the sentence BERT model. $m$ denotes the margin parameter and ensures that dissimilar pairs are separated by at least the margin distance $m$. The ReLU function is given by: $ReLU(x) = max(0,x)$. 

For \emph{Spider-Route}, we train 2 models. One for retrieving database schema with the question as the query with <question, database schema> as the sentence pair and the other for retrieving tables in a DB with the question as the query with <question, table schema> as the sentence pair. For \emph{Bird-Route} we train 2 models one for retrieving tables in a database, with relevant meta-data in context with the question as the query with <question, table schema+relevant domain statements> as the sentence pair and the other for retrieving domain statements in the meta-data of a database relevant to a question with <question, domain statement> as the sentence pair. For training, we use NVIDIA MIG A100 GPU with 10 GB RAM. We fine-tune the SBERT model, with 16 batch size, learning rate of $5e-6$, and an Adam optimizer for 2 epochs. It takes approximately 11 minutes to arrive at the best checkpoint.
%, when we observe the minimum validation loss.

%Since the schema of databases in Spider fit in the context length of BERT, we fine-tune the base model to directly rank the databases. The model applies the loss on a train set (construction of train set detailed in \ref{sec:spider-route}) with samples of <question, schema, label>.}\\
%\textcolor{red}{The BIRD-SQL dataset contains schemas with large number of tables and columns along with question specific evidences. When fine-tuning with evidences, we train 2 models. The first model ranks the evidences in a database given a question. It contains samples of <question, evidence, label>. The second model ranks the tables in a database based on relevance to a question. It contains samples of <question, evidence + table, label>. Finally, when fine-tuning without metadata, we train a model to directly rank the tables without evidences. It applies the loss function on samples of <question, table, label>. The train dataset creation is detailed in \ref{sec:bird-route}.}\\

The inference process used for identifying Top-K databases using the fine-tuned task-specific embeddings is the same as the one discussed in Section \ref{sec:pre}, except we use the fine-tuned sentence-BERT models as opposed to pre-trained models, to generate the embeddings.

\section{Results}

\subsection{Metric}
For each question, we calculate the recall@1 (R1), recall@3 (R3) and mean average precision (mAP). For a question, Recall@1 is 1  if the DB ranked the highest by an approach is the ground truth DB (i.e. the database on which the question is posed in the dataset) and is 0 otherwise. The value of Recall@3 is 1 if the ground truth DB is present amongst the top-3 ranked DBs. The mAP for a question is calculated by the formula $1/{i}$, where i is the index at which the correct DB is found among the ranked DBs, with the highest ranked DB is at index 1. 
We compute results within verticals and across verticals taking into consideration the mapping of DBs to vertical clusters (section \ref{sec:dataset}).
For a sample, if the clusters of the highest ranked DB and the ground truth DB are the same then the Recall@1 across vertical is 1, whereas within vertical is 0, indicating that the confusion about the most relevant database is within domain.
On the other hand, if the clusters are different then the Recall@1 across vertical is 0, whereas within vertical is 1, indicating the confusion is across verticals, which is even worse. If the correct DB gets predicted as the most relevant one then both within and across vertical Recall@1 is 1. 
Many DBs belong to singleton clusters and in such cases, it does not make sense to evaluate if top-k ranked DBs belong to the same cluster as the ground-truth DB cluster. Hence, for within vertical we calculate only the Recall@1. As each of the DBs in the $Te_{out}$ of \emph{Bird-Route} belong to distinct domains, we do not compute the results for within verticals and across verticals for this split.

\begin{table}[!t]
  \caption{Results on \emph{Spider-Route}  $Te_{out}$ (20 DBs). W-V: Within Vertical, A-V: Across Vertical, Emb.: Embedding, Rk: Recall@k, mAP: mean Average Precision}
  \label{tab:spider_cross}
  \small
  \begin{tabular}{l c c c c c c c c c}
    \toprule
          &\multicolumn{3}{c}{Overall} & \multicolumn{1}{c}{W-V}& \multicolumn{3}{c}{A-V}\\
    Model &R1&R3&mAP & R1 &R1&R3&mAP\\
    \midrule
    Llama3 &95.45&99.12&97.15&96.71&98.64&99.80&99.18\\
    Pretrained Emb. &87.71&99.35&92.61&89.94&96.61&99.22&97.80\\
    Task-spec. Emb.  &91.78&97.97&94.70&94.77&97.00&99.12&98.00\\
    \bottomrule
  \end{tabular}
\end{table}

\begin{table}[!t]
  \caption{Results on \emph{Spider-Route} $Te_{in}$ (140 DBs).}
  \label{tab:spider_in}
  \small
  \begin{tabular}{l c c c c c c c c c}
    \toprule
          &\multicolumn{3}{c}{Overall} & \multicolumn{1}{c}{W-V}& \multicolumn{3}{c}{A-V}\\
    Model &R1&R3&mAP & R1 &R1&R3&mAP\\
    \midrule
    Llama3 & 59.84&60.13&59.99&78.96&80.88&81.17&81.01\\
    Pretrained Emb. &44.09&67.82&54.42&68.87&75.21&86.16&80.08\\
    Task-spec. Emb.   &55.04&78.09&65.13&75.89& 79.15&90.10&84.21\\
    \bottomrule
  \end{tabular}
\end{table}

\begin{table}[!t]
  \caption{Results on \emph{Bird-Route}  $Te_{out}$ with metadata (11 DBs)}
  \label{tab:bird_cross}
  \begin{tabular}{l c c c}
    \toprule
    Model &R1&R3&mAP\\
    \midrule
    %Llama3 &86.04&88.07&86.96\\
    Pretrained Emb. &97.71&99.67&98.64\\
    Task-spec. Emb. &98.70&99.80&99.22\\
    \bottomrule
  \end{tabular}
\end{table}

\begin{table}[!ht]
  \caption{Results on \emph{Bird-Route} $Te_{in}$ with metadata (69 DBs)}
  \label{tab:bird_in}
  \small
  \begin{tabular}{l c c c c c c c c c}
    \toprule
          &\multicolumn{3}{c}{Overall} & \multicolumn{1}{c}{W-V}& \multicolumn{3}{c}{A-V}\\
    Model &R1&R3&mAP & R1 &R1&R3&mAP\\
    \midrule
    %Llama3& & & & & & & \\
    Pretrained Emb. &80.21&95.00&86.94&88.18&92.03&97.36&94.36\\
    Task-spec. Emb.  &89.05&95.87&92.87&94.79&95.25&97.90&95.91\\
    \bottomrule
  \end{tabular}
\end{table}

\subsection{Research Questions}\label{sec:rq}

\begin{table*}[!t]
  \caption{Results on  test split of ($Te_{in}$  + $Te_{out}$) \emph{Spider-Route}.  Reducing the databases from 160 to 20}
  \label{tab:spider_subset}
  %\small
  \begin{tabular}{l |c c c |c c c| c c c| c c c| c c c}
    \toprule
          &\multicolumn{3}{c|}{160 DB}&\multicolumn{3}{c|}{120 DB}&\multicolumn{3}{c|}{80 DB}&\multicolumn{3}{c|}{60 DB}&\multicolumn{3}{c}{20 DB}\\
    Model &R1&R3&mAP&R1&R3&mAP&R1&R3&mAP&R1&R3&mAP&R1&R3&mAP\\
    \midrule
    %Re-ranked Llama &&&&&&&&&&&&&&&\\
    Pretrained Emb. & 51.13&74.21&61.47&51.37&77.76&63.23&56.48&85.59&69.55&65.81&86.36&74.94&87.71&99.35&92.61\\
    Task-spec. Emb. & 60.38&80.48&69.40&63.47&84.41&73.28&63.51&88.10&80.30&66.18&90.72&87.42&91.78&97.97&94.70\\
    \bottomrule
  \end{tabular}
\end{table*}
\begin{table}[!t]
  \caption{Comparison of $T_{in}$ and $T_{out}$ of \emph{Spider-Route}. For $T_{in}$, averaged over 7 sets of 20 DBs sampled w/o replacement}
  \label{tab:spider_in_vs_cross_sampling}
  \begin{tabular}{l c c c  c c c}
    \toprule
          %&\multicolumn{7}{c}{In-domain}&\multicolumn{3}{c}{Cross-domain}\\
          &\multicolumn{3}{c}{In-domain}% & \multicolumn{1}{c}{Within vertical}& \multicolumn{3}{c}{Across vertical}
          &\multicolumn{3}{c}{Cross-domain}\\
    Model &R1&R3&mAP %& R1 &R1&R3&mAP
    &R1&R3&mAP\\
    \midrule
    Llama3&79.81&93.78&86.21%&85.54& 94.26&97.84&95.90 
    &95.45&99.12&97.15\\
    Pretrained Emb.  &63.14&89.93&75.32%&74.12&89.02&96.65 &92.67
    &87.71&99.35&92.61\\
    Task-spec. Emb.&72.65&93.35&82.03%&78.60&94.03&98.34&96.09 
    &91.78&97.97&94.70\\
    %Sampled DB Task-spec. Emb.&79.71&95.65&87.08&82.60&97.10&98.55&97.82&-&-&-\\
    \bottomrule
  \end{tabular}
\end{table}
\begin{table}[!t]
  \caption{Comparison of with and without meta-data for \emph{Bird-Route} cross-domain test set $Te_{out}$}
  \label{tab:bird_w_vs_wo_md_cross_dom}
  \begin{tabular}{l c c c c c c}
    \toprule
          &\multicolumn{3}{c}{With Metadata}&\multicolumn{3}{c}{Without Metadata}\\
    Model &R1&R3&mAP&R1&R3&mAP\\
    \midrule
    %Llama3 &86.04&88.07&86.96&87.09&88.33&87.67\\
    Pretrained Emb. &97.71&99.67&98.64&91.00&99.28&94.91\\
    Task-spec. Emb. &98.70&99.80&99.22& 95.17&99.41&97.21\\
    \bottomrule
  \end{tabular}
\end{table}

We try to address the following Research Questions (RQs):
\\
\textbf{RQ1:Does domain-specific (availability of questions from the same database) or cross-domain training assist to improve the performance of the DB-routing task?}  Table \ref{tab:spider_cross} and Table \ref{tab:spider_in} provides results for \emph{Spider-Route} for cross-domain $Te_{out}$ and in-domain $Te_{in}$ test sets. Similarly, Table \ref{tab:bird_cross} and Table \ref{tab:bird_in} provides results for \emph{Bird-Route} for cross-domain $Te_{out}$ and in-domain $Te_{in}$ test sets. Results of Task-specific embeddings are always better across all metrics than pre-trained embeddings for both in-domain as well cross-domain setting. As expected, the improvement in the results is always higher for in-domain setting ($\sim$25\% and $\sim$11\% jump in R@1 for \emph{Spider-Route} and \emph{Bird-Route}) as compared to cross-domain ($\sim$4.6\% and $\sim$1\% jump in R@1 for \emph{Spider-Route} and \emph{Bird-Route}). We observe that for \emph{Spider-Route} dataset Llama3 achieves much better performance as compared to the pre-trained embedding-based model. However, the results of task-specific embeddings generated by a smaller BERT model of only 110 Million parameters are more comparable to the results by a much larger Llama3 70 Billion parameter model. This validates the efficacy of the task-specific fine-tuning and synthesized training data used for the task. 

It can be observed that the cross-domain results are better than in-domain for all methods, which is counter-intuitive, specifically for the task-specific embedding-based approach, where the model is trained with in-domain DBs. However, this is solely because of the disparity in the number of DBs in the in-domain and cross-domain sets. The number of DBs in the in-domain split is much higher than that of the cross-domain split for both \emph{Spider-Route} and \emph{Bird-Route}. We perform an experiment where we sample 7 sets of 20 databases, each without replacement from the 140 databases of $Te_{in}$ of \emph{Spider-Route}. We sample such that the vertical cluster distribution of each set is similar to that of the vertical cluster distribution of the cross-domain split of \emph{Spider-Route} in terms of number of clusters and DBs per cluster. We compute results per set and take an average over all the sets. We compare these results of the in-domain split with the original cross-domain split (Table \ref{tab:spider_in_vs_cross_sampling}). We observe that the performance of the in-domain set with Llama3 and pre-trained embeddings is lower as compared to the performance of the cross-domain split. This demonstrates the overall difficulty of these in-domain sets. We observe the same trend for the performance of task-specific embedding-based approach. However, this approach reduces the difference in the performance of in-domain set and cross-domain split, from 24.57 for the pre-trained embedding-based approach to 19.13 for Recall@1, showcasing the advantage of in-domain training. The same trend can be observed for other metrics as well. We perform similar experiments for \emph{Bird-Route} and observe similar trends (Table \ref{tab:bird_in_vs_cross_sampling} in Appendix \ref{app:bird-expt}).
%There is a much higher number of DBs in the in-domain split as compared to the number of DBs in the cross-domain split

%\textbf{RQ2: Is in-domain performance better than cross-domain?}

\textbf{RQ2: Does similarity in domains of the data-sources affect the performance of DB Routing task?} As it can be observed in tables \ref{tab:spider_cross}, \ref{tab:spider_in}, \ref{tab:bird_in}, the across vertical Recall@1 is always better than within vertical Recall@1, for all the methods and dataset splits. This indicates that the ground truth DB is mostly confused with the DBs belonging to the same vertical cluster and thus similar domain, than the DBs which belong to distinct domains. Thus, more similar the domains more difficult the task of routing to the right DB becomes. This can be mainly because the pre-trained as well as learned embeddings of the DB schema within the same cluster lie closer in the embedding space and thus are difficult to distinguish. 
\\
\textbf{RQ3: Does an increase in the number of data sources decrease the performance of DB Routing Task?}
We combine the in-domain ($Te_{in}$) and cross-domain ($Te_{out}$) test splits of the datasets resulting in 160 DBs for \emph{Spider-Route} (Table \ref{tab:spider_subset}) and 80 DBs for \emph{Bird-Route} (Table \ref{tab:bird_subset} in the Appendix \ref{app:bird-expt}). We start with a randomly selected small number of DBs and check the performance for all the questions pertaining to those DBs. We keep adding randomly sampled DBs, until we have all the DBs in the repository, simulating the scenario of the availability of more number of data sources. We observe that for a lesser number of DBs the performance on all metrics is very high.As the number of DBs increases, the performance drops monotonically. However, the drop in performance is not consistent for every step. This is because when the DBs are sampled randomly, the vertical cluster distribution of the resulting set may not be consistent, varying the task difficulty over the steps.
% simulating the scenario of the availability of more number of data source to choose from to route the given question
\\
\textbf{RQ4: Does domain-specific external knowledge facilitate in improving the performance of DB Routing task?}
It can be observed in Table \ref{tab:bird_w_vs_wo_md_cross_dom}, the performance of \emph{Bird-Route} cross-domain test set $Te_{out}$, across all the metrics and methods is better when the retrieved relevant domain statements are augmented for the task of DB Routing (With Meta-data) as opposed to with no augmentation (without meta-data). This demonstrates the positive effect of augmentation of external domain knowledge for the DB Routing.

\textbf{Error Analysis:} We take into consideration the most difficult (having lowest performance) split for analyzing errors, which is the in-domain $Te_{in}$ split of \emph{Spider-Route}. We compare the erroneous samples of two approaches at a time (a) Llama3 Vs pre-trained embedding (b) Pre-trained  Vs task-specific embedding. The details of the analysis can be found in the Appendix \ref{app:err}. We observe that the approaches are not complementary and largely make similar mistakes. Majority of errors are due to question ambiguity and confusion between DB Schema belonging to the same vertical cluster. Sometimes the errors are due to partial question-schema match (mainly embedding-based approach) or inability to perform logical reasoning capabilities. We further find that task-specific fine-tuning often cannot rectify ambiguous questions but is able to differentiate between DBs from the same vertical better than the pre-trained model.
\section{Conclusion}
As part of this work, we define the novel task of routing an end-user query, posed as a part of an enterprise search, to an appropriate database as the data-source, which can correctly answer the query. We create baselines for our synthetically created datasets and demonstrate that open-source LLM performs better than embedding base approaches, but suffers from token length issues. Embedding-based approaches get benefited by task-specific fine-tuning, more when there is an availability of annotated data specific to the database domain as compared to cross-domain. We further observe that the task becomes more difficult (i) with an increase in the number of data-sources, (ii) having data-sources closer in terms of their domains, (iii) having databases with unavailability of external domain knowledge required to interpret its entities and (iv) with ambiguous and complex query having partial match with multiple database schema entities, requiring more fine-grained understanding of the data-sources or logical reasoning for routing it to an appropriate source. 

As part of the future work, we plan to develop a more sophisticated solution to better address the task. We plan to extend the dataset with (i) removal of ambiguous queries (ii) specifically designing queries that can be addressed by more than one data-source, and (iii) introducing queries that are unanswerable by all the data-sources. (iv) combining it with existing datasets of heterogeneous data-sources, other than databases, such as knowledge graphs, text documents, etc.
\section{Appendices}
\subsection{Schema Format: Data Definition Language}\label{app:ddl}
Here we demonstrate the format in which we include DB schema in the Language Model prompt. We use Database name followed by Data Definition Language (DDL) script  (`CREATE TABLE' commands) to define tables and columns as illustrated in Table \ref{tab:schema_format}.
\begin{table}[!ht]
  \caption{Schema Format: Data Definition Language (DDL)}
  \label{tab:schema_format}
  \small
  \begin{tabular}{ l }
    \toprule
    CREATE TABLE perpetrator (\\
    'perpetrator id' INTEGER PRIMARY KEY,\\
    'people id' INTEGER FOREIGN KEY,\\
    date TEXT,\\
    year INTEGER,\\
    location TEXT,\\
    country TEXT,\\
    killed INTEGER,\\
    injured INTEGER,\\
    );\\
    CREATE TABLE people (\\
    'people id' INTEGER PRIMARY KEY,\\
    name TEXT,\\
    height INTEGER,\\
    weight INTEGER,\\
    'home town' TEXT,\\
    );\\
    ...\\
    \bottomrule
  \end{tabular}
\end{table}

\subsection{Llama3 Prompt}\label{app:prompt}
Here in Table \ref{}, we provide the details of the prompt we use for the ranking the DBs for a given a question with Llama3 language model (LM).
\begin{table}[!ht]
  \caption{Llama 3 Prompt}
  \label{tab:prompt}
  \begin{tabular}{ l }
    \toprule
    You are a database administrator \\ and have designed the following databases\\ whose names and corresponding schema is given as:\\
    Database 1: <database name>>\\
    Database schema: <DDL Script of database schema>\\
    %Database Metadata: <List of domain statements>\\
    .\\
    .\\
    Database n: <database name>>\\
    Database schema: <DDL Script of database schema>\\
    %Database Metadata: <List of domain statements>\\
    \\
    Your task is to find the names of the 3 most relevant databases \\ to answer the given question correctly.\\ Your response must contain 3 relevant database names \\ in descending order of relevance in the given format: \\<database 1,database 2,database 3>.\\ The first database  must be most relevant to the question.\\ Only provide the 3 database names and not  any explanation.\\
    Question: <question> \\
    Top-3 Ranked Databases: \\
    \bottomrule
  \end{tabular}
\end{table}

\subsection{Vertical Clusters}\label{sec:verticals}
Each of the 11 DBs of $Te_{out}$ of \emph{Bird-Route} belong to a distinct cluster. The names of the DBs are: Thrombosis Prediction, California Schools, Card Games, Debit Card Specification, Toxicology, Financial, Codebase Community, European Football, Formula 1, Student Club, and Superhero. Here we illustrate the manually formulated clusters of verticals (domains) of 140 databases in  \emph{Spider-Route} in-domain Test split {$Te_{in}$} (Table \ref{tab:spider-in-domain_clusters}), 69 databases of \emph{Bird-Route} in-domain Test set {$Te_{in}$} (Table \ref{tab:bird-SQL}) and 20 databases of \emph{Spider-Route} cross-domain Test set{$Te_{out}$} (Table \ref{tab:spider_dev}).

\begin{table}%[!ht]
  \caption{Bird-SQL in-domain clusters(69 databases)}
  \label{tab:bird-SQL}
  \small
  \begin{tabular}{ l }
    \toprule
    \midrule    1. book\_publishing\_company,books,authors,citeseer\\
    2. cars\\
    3. college\_completion,computer\_student,cs\_semester,university,\\student\_loan\\
    4. movie,movie\_3,movie\_platform,movielens,movies\_4,music\_platform\_2,\\disney\\ 5. car\_retails,retail\_complains,retail\_world,retails,sales,regional\_sales,\\works\_cycles,shipping\\
    6. food\_inspection,food\_inspection\_2\\    7. olympics,professional\_basketball,soccer\_2016,shooting,hockey,\\european\_football\_1,ice\_hockey\_draft\\
    8. cookbook,menu\\
    9. craftbeer,beer\_factory\\    10. world,world\_development\_indicators,mondial\_geo,address,\\chicago\_crime,restaurant\\
    11. sales\_in\_weather,bike\_share\_1\\
    12. shakespeare\\
    13. law\_episode,simpson\_episodes\\
    14. software\_company\\
    15. airline\\
    16. app\_store\\
    17. synthea\\
    18. social\_media\\
    19. video\_games\\
    20. superstore\\
    21. public\_review\_platform\\
    22. image\_and\_language\\
    23. talkingdata\\
    24. genes\\
    25. music\_tracker\\
    26. codebase\_comments\\
    27. mental\_health\_survey\\
    28. donor\\
    29. legislator\\
    30. language\_corpus\\
    31. human\_resources\\
    32. coinmarketcap\\
    33. trains\\
    \bottomrule
  \end{tabular}
\end{table}

\begin{table}%[!h]
  \caption{Spider in-domain set clusters (140 databases)}
  \label{tab:spider-in-domain_clusters}
  \small
  \begin{tabular}{ l }
    \toprule
    \midrule    1. college\_2,college\_1,college\_3,csu\_1,dorm\_1,student\_1,\\
    2. student\_assessment,e\_learning,behavior\_monitoring\\
    3. flight\_1,flight\_4,flight\_company,aircraft,pilot\_record\\
    4. music\_2,music\_1,music\_4,musical,sakila\_1,chinook\_1\\
    5. school\_finance,school\_bus\\
    6. coffee\_shop,restaurant\_1\\    
    7. voter\_2,election,candidate\_poll,election\_representative,\\county\_public\_safety\\
    8. driving\_school,bike\_1\\    9. baseball\_1,game\_1,university\_basketball,race\_track,\\sports\_competition,gymnast,school\_player,swimming,formula\_1,\\match\_season,climbing,riding\_club,game\_injury,soccer\_2,soccer\_1\\    10. small\_bank\_1,cre\_Docs\_and\_Epenses,customers\_and\_invoices,\\customers\_card\_transactions,insurance\_policies,insurance\_fnol, \\insurance\_and\_eClaims,tracking\_share\_transactions,loan\_1,\\tracking\_orders,customer\_complaints,product\_catalog,\\products\_gen\_characteristics,customers\_and\_addresses,\\customers\_campaigns\_ecommerce,products\_for\_hire,\\11. customers\_and\_products\_contacts,customer\_deliveries,\\document\_management,cre\_Doc\_Tracking\_DB,cre\_Doc\_Control\\\_Systems,tracking\_grants\_for\_research,\\cre\_Drama\_Workshop\_Groups,solvency\_ii\\
    12. movie\_1,cinema,film\_rank\\
    13. local\_govt\_mdm,e\_government,local\_govt\_and\_lot,\\local\_govt\_in\_alabama\\
    14. workshop\_paper,scientist\_1,journal\_committee,icfp\_1\\    15. company\_office,gas\_company,culture\_company,\\company\_employee,company\_1,hospital\_1,hr\_1,\\department\_management,store\_1,department\_store,store\_product,\\shop\_membership,manufactory\_1,manufacturer,entrepreneur\\
    16. storm\_record,station\_weather\\
    17. hone\_1,phone\_market,device\\
    18. assets\_maintenance,machine\_repair\\
    19.ship\_1,ship\_mission\\
    20. partment\_rentals,inn\_1,cre\_Theme\_park\\
    21. club\_1,activity\_1,decoration\_competition\\
    22. body\_builder,wrestler\\
    23.train\_station,railway\\
    24.party\_host,party\_people\\
    25.network\_2,twitter\_1\\
    26.allergy\_1\\
    27.medicine\_enzyme\_interaction\\
    28.farm\\
    29.wine\_1\\
    30.debate\\
    31. architecture\\
    32. epinions\_1\\
    33. city\_record\\
    34. news\_report\\
    35. entertainment\_awards\\
    36. tracking\_software\_problems\\
    37. perpetrator\\
    38. book\_2\\
    39. wedding\\
    40. mountain\_photos\\
    41. roller\_coaster\\
    42. browser\_web\\
    43. protein\_institute\\
    44. performance\_attendance\\
    45. program\_share\\
    46. theme\_gallery\\
    \bottomrule
  \end{tabular}
\end{table}

\begin{table}[!ht]
  \caption{Clusters of verticals for 20 databases in \emph{Spider-Route} cross-domain Test set ($Te_{out}$)  }
  \label{tab:spider_dev}
  \begin{tabular}{ l }
    \toprule
    1. concert\_singer, singer, orchestra\\
    2. pets\_1,  dog\_kennels, real\_estate\_properties, \\employee\_hire\_evaluation\\
    3. course\_teach, student\_transcripts\_tracking, network\_1, \\museum\_visit\\
    4. voter\_1, world\_1, car\_1, wta\_1, poker\_player, battle\_death\\
    5. cre\_Doc\_Template\_Mgt\\
    6. tvshow\\
    7. flight\_2\\
    \bottomrule
  \end{tabular}
\end{table}

\subsection{Bird-Route Experiments}\label{app:bird-expt}
Here, we provide the results of experiments with the \emph{Bird-Route} dataset. Table \ref{tab:bird_in_vs_cross_sampling} illustrates the comparison of In-domain $Te_{in}$  Cross-domain $Te_{out}$ splits for \emph{Bird-Route}. For a fair comparison with the cross-domain split consisting of 11 DBs,  we form 6 sets of  11 DBs each from $Te_{in}$ of \emph{Bird-Route}, where DBs for each set are sampled without replacement. Then, we take the average of the results over these sets. This Table is referred to in section \ref{sec:rq} for addressing research question \textbf{RQ1}. Table \ref{tab:bird_subset} illustrated the results on \emph{Bird-Route} when we increase the number of DBs from 20 to 80 to address the research question \textbf{RQ3} in section \ref{sec:rq}. 
\begin{table}%[!h]
  \caption{Comparison of In-domain $Te_{in}$  Cross-domain $Te_{out}$ splits for \emph{Bird-Route}. In-domain results are averaged over the 6 sets, where each set consists of 11 databases randomly sampled from the $Te_{in}$ without replacement}
  \label{tab:bird_in_vs_cross_sampling}
  \begin{tabular}{l c c c c c c}
    \toprule
          &\multicolumn{3}{c}{In-domain}&\multicolumn{3}{c}{Cross-domain}\\
    Model &R1&R3&mAP&R1&R3&mAP\\
    \midrule
    %Re-ranked Llama 70b& & & &86.04&88.07&86.96\\
    Pretrained Emb &90.95&99.36&94.93&97.71&99.67&98.64\\
    Task-spec. Emb.&94.91&99.44&97.06&98.70&99.80&99.22\\
    \bottomrule
  \end{tabular}
\end{table}

\begin{table*}%[!ht]
  \caption{Results on  test split ( $Te_{in}$  + $Te_{out}$) of \emph{Bird-Route} with metadata.  Reducing the number of databases from 80 to 20}
  \label{tab:bird_subset}
  %\small
  \begin{tabular}{l |c c c |c c c |c c c}
    \toprule
          &\multicolumn{3}{c|}{80 DB}&\multicolumn{3}{c|}{50 DB}&\multicolumn{3}{c}{20 DB}\\
    Model &R1&R3&mAP&R1&R3&mAP&R1&R3&mAP\\
    \midrule
    %Re-ranked Llama &&&&&&&&&\\
    Pretrained Emb. &  86.26&95.92&90.74&86.53&96.48&91.16&96.26&99.28&97.68\\
    Task-spec. Emb. & 91.63&96.78&93.98&93.47&98.27&95.61&97.20&99.46&98.28\\
    \bottomrule
  \end{tabular}
\end{table*}

\subsection{Detailed Error Analysis}\label{app:err}
As discussed in section \ref{sec:rq},  we take into consideration the most difficult (having lowest performance) split for analyzing errors, which is in-domain $Te_{in}$ split of \emph{Spider-Route}. We compare the erroneous samples of two approaches at a time (a) Llama3 Vs pre-trained embedding (b) Pre-trained embedding Vs task-specific embedding. In this section, we elaborate on the error categories providing examples.

\subsubsection{Llama3 vs Pre-trained embedding }
We find that there are 217 samples where Llama3 is correct but pre-trained embeddings lead to incorrect DBs, 53 samples where pre-trained embeddings are correct but Llama3 is incorrect and 364 samples out of 1033 where both approaches go wrong. We see that while Llama3 performs better than the pre-trained embedding-based approach, the majority of the erroneous samples are common between these approaches. Thus the approaches are not completely complementary to each other. We sample 100 questions from a total of 634 erroneous ones to do an in-depth error analysis. %\textcolor{red}{This sample follows the overall distribution of the erroneous sample over the above defined categories} \todo[]{Validate this statement}. 

We find following the majority of errors can be categorized into the following four types of errors: 
\\
1. Question Ambiguity (36 out of 100): Since the original Spider dataset is designed for DB-specific NL2SQL tasks, there are many questions with incomplete context of the DB it is referring to.  Such questions may be answered by multiple DB schemas. For example,  the question `Find the number of albums.', can be answered by multiple databases in the vertical cluster related to music, viz. 'music\_2' and `chinook\_1'. We call these as \emph{ambiguous questions}. Another type of ambiguity exists when there can be multiple meanings of the same question. For example, in the question `Count the number of tracks.', tracks can mean both songs in a playlist as well as race tracks. Thus, the correct DB can be `race\_track' or `chinook\_1'.  Such questions are difficult for even a human to answer as expected. 
\\
2. Confusion between DB schemas belonging to the same Vertical cluster (34 out of 100): Both approaches find it confusing to differentiate between two DBs whose schema are semantically similar, but only one of them has the required columns/tables to answer the question. While this type of error is expected from the pre-trained embedding-based model, we find that even Llama3 performs this type of error. For example, the question `What is the name and distance for aircraft with id 12?' belongs to the DB 'flight\_1'. The approaches predict `aircraft' or `flight\_company' as the top-1 databases. Both these databases are semantically overlapping with columns for aircraft names, IDs, and other information. However, none of them have any information related to the distance covered. We observe that for samples with this type of error, the correct DB lies among the top 3 databases. To resolve this type of error there is a need for fine-grained retrieval based on more data such as column names or column values, as opposed to the match based on the complete schema. %What these models lack is a backtracking mechanism that lets them check the next databases if the highest-ranked database does not answer the question.
\\
%For some of the questions where both models predicted wrong, the SBERT model chose a database due to a partial match with the question while ignoring the other important information. 
3. Partial question schema match (25 out of 100): This type of error is common for the pre-trained embedding-based approach. For the question, `Find the number of different departments in each school whose number of different departments is less than 5.', the pre-trained embedding-based approach ranks 'department\_management' DB at the top as opposed to 'college\_1'. 'department\_management' has more mentions of the  'Department' entity making the model biased towards this DB. However, Llama3 does not suffer from this problem. Thus, this demonstrates that schema-level representations are not able to capture the required fine-grained DB semantics properly.
%The database Llama predicted in this case were wrong due to ambiguity of the question itself. Thus, for this task we find that certain columns of a database are significantly more important than the rest to answer a certain question and schema level representations are not able to capture the required semantics properly.
\\
%In category (ii), a large number of errors (x out of 100) \todo[]{insert number} are due to ambiguous questions. It is only by chance that SBERT could classify these questions correctly. Other than that,
4. Approaches lacking logical reasoning capability (5 out of 100) :
We find Llama3 sometimes lacks logical reasoning capability. For example, for the question `What are the types of every competition and in which countries are they located?', Llama predicts 'formula\_1' instead of 'sports\_competition' as top-1 database. Whereas, 'formula\_1' is a single competition and 'sports\_competition' has information about multiple competitions. A human can easily perform this type of reasoning, which Llama3 finds difficult. 

%Finally, in category (i), while some questions are ambiguous, they are only a minority. SBERT mostly gets confused between databases of the same cluster which are semantically very similar to the correct database. In most instances, the correct database lies among the top 3 predicted databases. Like in category (iii), here too SBERT predicts databases where there is a partial with the question in addition to logical errors. For example, `Give me the start station and end station for the trips with the three oldest id.' is the given question, to which SBERT predicts `flight\_4'. Here, `start' and `end' cause a partial match with `destination' and `source' airport. We see that while both the models make these errors, SBERT is more prone to make such errors based only on textual similarity as its number of errors (217) are significantly more than that of Llama (53).

\subsubsection{Fine-tuned vs Pre-trained embedding}
There are 179, 65 and 403 samples where the pre-trained embedding-based approach goes wrong but fine-tuned embeddings yield correct prediction, fine-tuned embeddings introduce errors to correctly predicted samples by pre-trained embeddings, and both approaches fail to predict the correct DB at the top, respectively. We take a sample of 100 questions to analyze the improvements through fine-tuning. For 58 samples out of 100 where both approaches fail, we find fine-tuned embedding-based model makes similar errors as discussed in the prior section.  

The remaining 32 samples are where the errors made by the pre-trained embedding-based approach are rectified by the task-specific embeddings, whereas for the remaining 10 samples the task-specific fine-tuning negatively affects the performance. We find that the majority (6 out of 10) of errors made by only task-specific fine-tuning approaches are due to question ambiguity. Thus, we find that the task-specific fine-tuning mainly helps to reduce the confusion between DB schemas and is able to differentiate between DBs from the same vertical better than the pre-trained model (13 out of 32), and errors due to partial match (5 out of 32), sometimes it also corrects the errors made by pre-training embedding due to question ambiguity (11 out of 32).

\bibliographystyle{ACM-Reference-Format}
\bibliography{sample-base}

\begin{comment}

\appendix

\section{Research Methods}

\subsection{Part One}

Lorem ipsum dolor sit amet, consectetur adipiscing elit. Morbi
malesuada, quam in pulvinar varius, metus nunc fermentum urna, id
sollicitudin purus odio sit amet enim. Aliquam ullamcorper eu ipsum
vel mollis. Curabitur quis dictum nisl. Phasellus vel semper risus, et
lacinia dolor. Integer ultricies commodo sem nec semper.

\subsection{Part Two}

Etiam commodo feugiat nisl pulvinar pellentesque. Etiam auctor sodales
ligula, non varius nibh pulvinar semper. Suspendisse nec lectus non
ipsum convallis congue hendrerit vitae sapien. Donec at laoreet
eros. Vivamus non purus placerat, scelerisque diam eu, cursus
ante. Etiam aliquam tortor auctor efficitur mattis.

\section{Online Resources}

Nam id fermentum dui. Suspendisse sagittis tortor a nulla mollis, in
pulvinar ex pretium. Sed interdum orci quis metus euismod, et sagittis
enim maximus. Vestibulum gravida massa ut felis suscipit
congue. Quisque mattis elit a risus ultrices commodo venenatis eget
dui. Etiam sagittis eleifend elementum.

Nam interdum magna at lectus dignissim, ac dignissim lorem
rhoncus. Maecenas eu arcu ac neque placerat aliquam. Nunc pulvinar
massa et mattis lacinia.
\end{comment}
\end{document}